# A Homogeneous Ensemble of Artificial Neural Networks for Time Series Forecasting

Ratnadip Adhikari
School of Computer & Systems Sciences
Jawaharlal Nehru University
New Delhi, India

R. K. Agrawal
School of Computer & Systems Sciences
Jawaharlal Nehru University
New Delhi, India

## ABSTRACT
Enhancing the robustness and accuracy of time series forecasting models is an active area of research. Recently, Artificial Neural Networks (ANNs) have found extensive applications in many practical forecasting problems. However, the standard backpropagation ANN training algorithm has some critical issues, e.g. it has a slow convergence rate and often converges to a local minimum, the complex pattern of error surfaces, lack of proper training parameters selection methods, etc. To overcome these drawbacks, various improved training methods have been developed in literature; but, still none of them can be guaranteed as the best for all problems. In this paper, we propose a novel weighted ensemble scheme which intelligently combines multiple training algorithms to increase the ANN forecast accuracies. The weight for each training algorithm is determined from the performance of the corresponding ANN model on the validation dataset. Experimental results on four important time series depicts that our proposed technique reduces the mentioned shortcomings of individual ANN training algorithms to a great extent. Also it achieves significantly better forecast accuracies than two other popular statistical models.

## General Terms
Time Series Forecasting, Artificial Neural Network, Ensemble Technique, Backpropagation.

## Keywords
Time Series Forecasting, Artificial Neural Network, Ensemble, Backpropagation, Training Algorithm, ARIMA, Support Vector Machine.

## 1. INTRODUCTION
Analysis and forecasting of time series is of fundamental importance in many practical domains. In time series forecasting, the historical observations are carefully studied to build up a proper model which is then used to forecast unseen future values [1]. Over the years, various linear and nonlinear forecasting models have been developed in literature [1–3]. During the last two decades, Artificial Neural Networks (ANNs) have been widely used as attractive and effective alternative tools for time series modeling and forecasting [2,4]. Originally motivated by the intelligent neural structure of human brains, ANNs have gradually found extensive applications in solving a broad range of nonlinear problems and have drawn increasing attentions of research community. Their most distinguishing feature is the nonlinear, nonparametric, data-driven and self-adaptive nature [2,4,5]. ANNs do not require any *a priori* knowledge of the associated statistical data distribution process. They adaptively construct the appropriate model from only the raw data, learn from training experiences, and then intelligently generalize the acquired knowledge to predict the nature of unseen future events. Due to these outstanding properties, they have become a favorite choice for time series researchers. At present, various ANN-based forecasting techniques exist in literature. Some excellent reviews on recent trends and developments in ANN forecasting methodology can be found in the works of Zhang et al. [4], Kamruzzaman et al. [5] and Adya and Collopy [6].

The performance of an ANN model is very sensitive to the proper selection of network architecture, training algorithm, the number of hidden layers, the number of nodes in each layer, the proper activation functions, the significant time lags, etc. [2,4]. The selection of a suitable network training algorithm is perhaps the most critical task in ANN modeling. So far, the classic *backpropagation*, developed by Rumelhart et al. [7] is the best-known training method. It updates the network weights and biases in the direction of the most rapid decrease of the error function, i.e. negative of the gradient; hence, backpropagation is also known as the *gradient steepest descent* method [4,5,7]. Despite its simplicity and popularity, this algorithm suffers from a number of drawbacks, which are listed here [4,5]:

- It has a very slow convergence rate and so requires a lot of computational time for large-scale problems.
- There exists no robust technique for the optimal selection of the corresponding training parameters.
- The error surface of the standard backpropagation algorithm has a very complex pattern.
- Often, the backpropagation algorithm gets stuck at the local minimum solution instead of the desired global one.

To get rid of these weaknesses, several improved or modified versions of backpropagation have been proposed in literature. Some important among them include the Levenberg-Marquardt (LM) [8], Resilient Propagation (RP) [9], Scaled Conjugate Gradient (SCG) [10], One Step Secant (OSS) [11], and Broyden-Fletcher-Goldfarb-Shanno (BFGS) quasi-Newton [12] algorithms. Recently, Particle Swarm Optimization (PSO) [13,14] has also received considerable attentions in this area; e.g. Jha et al. [14] has effectively used two PSO-based training algorithms (viz. PSO-Trelea1 and PSO-Trelea2) for predicting a financial time series. Although, the modified algorithms have improved the performance of backpropagation training on many occasions [15,16], they could not overcome all its drawbacks.

For example, at present no algorithm can unconditionally guarantee a global optimal solution. Moreover, there is no straightforward way of selecting the best training algorithm specific to a particular problem [4].

In this paper, we propose a weighted ensemble scheme to combine multiple training algorithms. It is a well-known fact that combining forecasts improves the overall accuracy much better than the individual methods [17]. Seven different backpropagation techniques, which are mentioned above are considered for building our ensemble. The weight assigned to each of them is inversely proportional to the forecast errors obtained by the corresponding ANN on validation dataset. The final forecast of this combined ANN model is calculated as the weighted arithmetic mean of the forecasts obtained from individual training algorithms. To evaluate and compare the performance of our ensemble technique, we consider four real-world time series and two other popular statistical models, viz. Autoregressive Integrated Moving Average (ARIMA) and Support Vector Machine (SVM). The forecast errors of all the models are evaluated in terms of Mean Squared Error (MSE) and Mean Absolute Percentage Error (MAPE).

The rest of the paper is organized as follows. Section 2 describes the ANN methodology for time series forecasting and various network training algorithms. Our proposed ensemble scheme is explained in Section 3. Section 4 describes two popular statistical time series forecasting models. Obtained experimental forecast results and model comparisons are reported in Section 5. Finally, Section 6 concludes our paper.

## 2. ARTIFICIAL NEURAL NETWORKS

The most widely used ANNs for time series forecasting are Multilayer Perceptrons (MLPs) [2–5]. These are characterized by the feedforward architecture of an input layer, one or more hidden layers, and an output layer. The nodes in each layer are connected to those in the immediate next layer by acyclic links. In practical applications, it is enough to consider a single hidden layer structure [2,4,5].

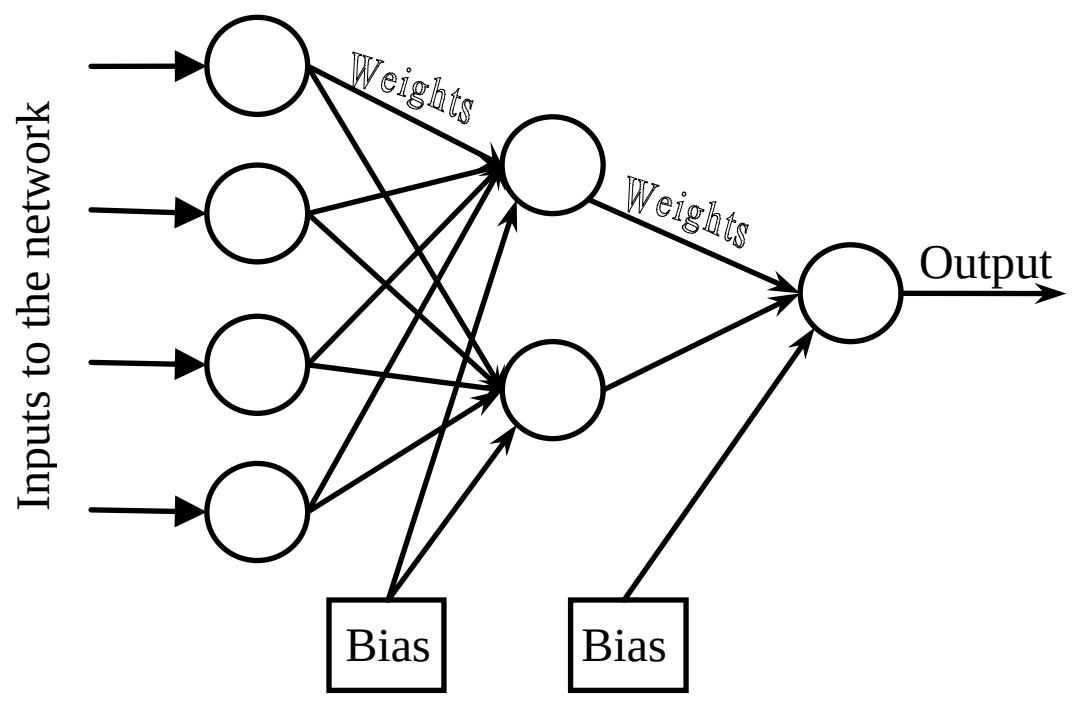

**Fig 1: A feedforward ANN architecture**

The output of an MLP with $p$ input and $h$ hidden nodes is expressed as [2,4]:

$$y_t = G\left(\alpha_0 + \sum_{j=1}^{h}\alpha_j F\left(\beta_{0j} + \sum_{i=1}^{p}\beta_{ij} y_{t-i}\right)\right). \quad (1)$$

Here, $y_{t-i}\,(i=1,2,\ldots,p)$ are the network inputs; $\alpha_{ij}, \beta_{jk}$ are the connection weights $(i=1,2,\ldots,p;\ j=1,2,\ldots,h)$; $\alpha_0, \beta_{0j}$ are the bias terms, and $F, G$ are respectively the hidden and output layer activation functions. Normally, logistic and identity functions are respectively used for $F$ and $G$, i.e. $F(x)=1/(1+\exp(-x))$ and $G(x)=x$. The model, given by the expression (1) is commonly referred as a $(p, h, 1)$ ANN model [14].

### 2.1 Backpropagation Training

Training is the iterative process for determining optimal network weights and biases. In this phase, the ANN model gradually learns from successive input patterns and target values and accordingly modifies the weights and biases. In a time series forecasting problem with training dataset $\{y_1, y_2, \ldots, y_N\}$, a $(p, h, 1)$ ANN model consists of $(N\text{-}p)$ training patterns with input vectors $\mathbf{Y}_i = \left[y_i, y_{i+1}, \ldots, y_{i+p-1}\right]^T$ and targets $y_{i+p}$, $(i=1, 2,\ldots, N\text{-}p)$. The backpropagation is a supervised training algorithm in which network weights and bias updating is carried out through the minimization of the error function [4,,5]:

$$E = \frac{1}{2}\sum_{t=p+1}^{N}\left(y_t - \hat{y}_t\right)^2 \quad (2)$$

where $y_t$ is the network output, calculated by Eq. (1) and $\hat{y}_t$ is the corresponding target. The algorithm starts with an initial vector $\mathbf{w}_0$ of weights and biases which is updated at each step (epoch) $i$ according to the gradient descent rule:

$$\left.\begin{aligned} \Delta\mathbf{w}_i &= -\eta\nabla E(\mathbf{w})\big|_{\mathbf{w}=\mathbf{w}_i} + \alpha\Delta\mathbf{w}_{i-1} \\ \mathbf{w}_i &= \mathbf{w}_i + \Delta\mathbf{w}_{i-1} \end{aligned}\right\}. \quad (3)$$

Here, $\eta$ and $\alpha$ are the learning rate and momentum factor respectively. The training process continues until some predefined minimum error or maximum number of epochs is reached. The obtained final values of $\mathbf{w}_i$ are used for all future predictions.

Various classes of improved backpropagation technique have been developed in literature. These improvements are done from different perspectives, such as robustness, direction of weights and bias updates, convergence rate, nature of error surface, local and global minima, etc. Here, we briefly discuss seven important network training algorithms, used in literature.

#### *2.1.1 The RPROP Algorithm*

In the steepest descent method, often the gradient achieves a very small value, causing negligible changes in weights and biases, even though these are actually far from their optimal values. To remove this drawback, the RPROP training algorithm was suggested. It considers only the signs of partial derivatives of the error functions to determine the directions of weight changes [9,18]. The magnitudes of derivatives have no effect on weight updates. This algorithm is very efficient and simple to apply.

#### *2.1.2 The Conjugate Gradient Algorithms*

Although the steepest descent direction provides the fastest decrease of the performance function but it does not necessarily means fastest convergence. Due to this fact, in conjugate gradient methods, a search is performed in conjugate directions of the gradient. An efficient method of this kind is the Scaled Conjugate Gradient (SCG) algorithm, developed by Moller [10]. It is used in this paper as one of the constituent training algorithm.

### 2.1.3 The Quasi-Newton Algorithms

This class of methods uses second order derivatives for weights and bias modifications. The optimum search direction is computed through $-\mathbf{H}^{-1}\nabla E(\mathbf{w})$, where **H** is the Hessian matrix of the error function. However, due to the expensive computational demand, the Hessian matrix is not calculated directly; rather it is assumed to be a function of the gradient which is iteratively approximated [18]. The most successful and widely applied quasi Newton method in literature is the Broyden-Fletcher-Goldfarb-Shanno (BFGS) algorithm [12,18]. Two others of this kind, used in the present paper are: the Levenberg-Marquardt (LM) [8] and One Step Secant (OSS) [11] algorithms. In particular, LM is so far the fastest training method in terms of convergence rate; however, it requires enormous amount of storage memory and mathematical computations [18].

### 2.1.4 PSO-Based Training Algorithms

PSO is an evolutionary optimization technique which is originally inspired from the intelligent working paradigm in birds flocks and fish schools [13,14]. From different variations of the basic PSO algorithm, in this paper we use PSO-Trelea1 and PSO-Trelea2, suggested by I. Trelea [19]. Consider a ($p$, $h$, $q$) neural network structure with $N$ particles in the swarm. Each particle represents an individual ANN structure with dimension $D=h(p+q+1)+q$, the total number of network parameters. The PSO algorithm begins by assigning randomized positions and velocities to each particle. The particles are moved through the $D$-dimensional search space until some error minimization criterion is satisfied. Every particle evaluates a fitness function (error function in this study) for it. The movements of the particles are governed by two best positions, viz. the personal and global, which are respectively the current personal best fitness of each particle and the current overall best fitness achieved across the whole swarm. Using these two values, the position and velocity for the $d^{th}$ dimension of the $i^{th}$ particle is updated as follows:

$$\left.\begin{aligned} v_{id}(t+1) &= a v_{id}(t) + b\left(p_d - x_{id}(t)\right) \\ x_{id}(t+1) &= x_{id}(t) + v_{id}(t+1) \\ p_d &= \frac{b_1 p_{id} + b_2 p_{gd}}{b_1 + b_2};\ b = \frac{b_1 + b_2}{2} \end{aligned}\right\}. \qquad (4)$$

Here, $x_{id}$, $v_{id}$, and $p_{id}$ are respectively the position, velocity and personal best position of the $i^{th}$ particle at the $d^{th}$ dimension; $p_{gd}$ is the global best position, obtained at the $d^{th}$ dimension and $a$, $b$ are two tuning parameters, which have two sets of values found for PSO-Trelea1 ($a$=0.6, $b$=1.7) and PSO-Trelea2 ($a$=0.729, $b$=1.494). In practice, 24 to 30 swarm particles are considered [14,19].

## 3. PROPOSED ENSEMBLE TECHNIQUE

Consider a time series $Y=\{y_1, y_2, \ldots, y_N\}$, which is divided into three subsets, viz. *validation*, *training*, and *testing*. These are used for selecting the best forecasting model, estimating the model parameters and assessing the out-of-sample forecast accuracy of the fitted model, respectively. Let $L_1, L_2, \ldots, L_k$ are $k$ (preferably an odd number) training algorithms to be used after determining the proper ANN structure for the time series. Our ensemble approach is now described below.

The determined ANN model is trained with every $L_i$ on the training dataset. These $k$ trained ANNs are then used to forecast the validation set and their subsequent forecast performances are measured using three error statistics, defined as:

$$\left.\begin{aligned} &\text{Mean Absolute Error (MAE)}=\frac{1}{n}\sum_{t=1}^{n}\left|y_t-\hat{y}_t\right| \\ &\text{Mean Squared Error (MSE)}=\frac{1}{n}\sum_{t=1}^{n}\left(y_t-\hat{y}_t\right)^2 \\ &\text{Mean Absolute Percentage Error (MAPE)}=\frac{1}{n}\sum_{t=1}^{n}\left|\frac{y_t-\hat{y}_t}{y_t}\right|\times 100 \end{aligned}\right\}$$

where, $y_t$ and $\hat{y}_t$ are the actual and forecasted observations, respectively and $n$ is the size of the forecasted dataset. Now, the weight for each training algorithm $L_i$ is computed as:

$$\left.\begin{aligned} w_i &= \exp(g_i), \\ g_i &= \frac{1}{\text{MAE}_i + \text{MSE}_i + \text{MAPE}_i} \end{aligned}\right\} \qquad (5)$$

where, the terms in the denominator of $g_i$ refer to the errors obtained by the ANN model, trained with the training algorithm $L_i$ on the validation dataset.

Equation (5) ensures that the weight assigned to a training algorithm is inversely proportional to its combined error (i.e. the sum of MSE, MAE, MAPE) on the validation set; so, the more error, the less weight and vice versa. The exponential function in weight calculation is used as it will regularize the effect of different magnitudes of the three error statistics.

After calculating the weights, each of the $k$ ANN models are individually trained on the whole in-sample dataset (i.e. training and validation sets combined) and their out-of-sample forecast for the test set is recorded. Let $\mathbf{D}_i$ be the vector of out-of-sample forecast values obtained by the ANN model, trained with $L_i$; the dimension of $\mathbf{D}_i$ ($i$=1, 2,…, $k$) is equal to the size of the test set. The final forecast vector **D** produced by our ensemble technique is the weighted arithmetic mean of all the $k$ ANN forecasts, i.e.

$$\mathbf{D} = \frac{w_1\mathbf{D}_1 + w_2\mathbf{D}_2 + \ldots + w_k\mathbf{D}_k}{w_1 + w_2 + \ldots + w_k} = \left(\sum_{i=1}^{k} w_i\mathbf{D}_i\right)\Bigg/\left(\sum_{i=1}^{k} w_i\right). \qquad (6)$$

Our proposed technique is a *homogeneous* ensemble, as it uses the same model (ANN) with different methods (i.e. training algorithms). The necessary steps in the proposed ensemble scheme are outlined below:

Algorithm: Ensemble of Multiple ANN Training Methods

1. Divide the time series into appropriate *validation*, *training*, and *testing* sets.
2. Select $k$ (preferably odd) training algorithms $L_1, L_2, \ldots, L_k$.
3. Determine the proper ANN model and network parameters using training and validation datasets.
4. Train the ANN with each $L_i$ to forecast the validation observations.
5. Calculate the weight $w_i$ for each $L_i$ by Eq. (5).

6. Train the ANN model with each $L_i$ on the whole in-sample dataset (i.e. training and validation sets combined).
7. Use each trained ANN to forecast the testing dataset and record the individual forecast vectors $\mathbf{D}_i$ ($i$=1, 2,…, $k$).
8. Calculate the combined forecasts using Eq. (6).

## 4. TWO STATISTICAL MODELS

In order to compare the forecast accuracy of our proposed method, two other well-known models, viz. Autoregressive Integrated Moving Average (ARIMA) and Support Vector Machine (SVM) are used in this paper. The present section gives a brief description about these two models.

### 4.1 ARIMA Models

The ARIMA or Box–Jenkins models [1] are based on the assumption that the observations of a time series are generated from a linear function of the past values and a random noise process [1,2,4]. These are actually a generalization of the Autoregressive Moving Average (ARMA) models [1] to deal with nonstationary time series. Mathematically, an ARIMA ($p$, $d$, $q$) model can be represented as follows:

$$\left.\begin{aligned} &\phi(L)(1-L)^d y_t = \theta(L)\varepsilon_t \\ &\text{where,} \\ &\quad \phi(L) = 1 - \sum_{i=1}^{p} \phi_i L^i \\ &\quad \theta(L) = 1 + \sum_{j=1}^{q} \theta_j L^j \\ &\text{and} \quad L y_t = y_{t-1} \end{aligned}\right\}. \tag{7}$$

Here, $p$, $d$, $q$ are the orders of the model, which refers to the autoregressive, degree of differencing and moving average processes, respectively; $y_t$ is actual time series and $\varepsilon_t$ is a random noise process; $\varphi(L)$ and $\theta(L)$ are lagged polynomials of orders $p$, $q$ with coefficients $\varphi_i$, $\theta_i$ ($i$=1, 2,…, $p$; $j$=1, 2,…, $q$), respectively and $L$ is the lag (or backshift) operator. This model transforms a nonstaionary time series to a stationary one by successively ($d$ times) differencing it. Usually, a single differencing is sufficient for most practical time series. The suitable ARIMA model is estimated through the famous Box-Jenkins methodology, which includes three iterative steps, viz. *model building*, *parameter estimation*, and *diagnostic checking* [1,2]. For seasonal time series forecasting, a variation of the basic ARIMA model, commonly known as the SARIMA$(p,d,q)\times(P,D,Q)^s$ model ($s$ is the seasonal period) was developed by Box and Jenkins [1], which is also used in this paper.

### 4.2 SVM Model

SVM is a new statistical learning theory, developed by Vapnik and co-workers at the AT & T Bell laboratories in 1995 [21]. It is based on the Structural Risk Minimization (SRM) principle and its aim is to find a decision rule with good generalization ability through selecting some special data points, known as *support vectors* [21,22]. Time series forecasting is a branch of Support Vector Regression (SVR) in which an optimal separating hyperplane is constructed to correctly classify real-valued outputs. Given a training dataset of $N$ points $\{\mathbf{x}_i, y_i\}_{i=1}^{N}$ with $\mathbf{x}_i \in \mathbb{R}^n$, $y_i \in \mathbb{R}$, SVM attempts to approximate the unknown data generation function in the following form: $f(x)=\mathbf{w}\cdot\varphi(\mathbf{x})+b$, where $\mathbf{w}$ is the weight vector, $\varphi$ is the nonlinear mapping to a higher dimensional feature space and $b$ is the bias term. Using the Vapnik's $\varepsilon$-insensitive loss function [21,22], the SVM regression is converted to a Quadratic Programming Problem (QPP) to minimize the empirical risk:

$$J\left(\mathbf{w}, \xi_i, \xi_i^*\right) = \frac{1}{2}\|\mathbf{w}\|^2 + C\sum_{i=1}^{N}\left(\xi_i + \xi_i^*\right) \tag{8}$$

where, C is the positive regularization constant and $\xi_i$, $\xi_i^*$ are the positive slack variables. After solving the associated QPP, the optimal decision hyperplane is given by:

$$y(\mathbf{x}) = \sum_{i=1}^{N_s}\left(\alpha_i - \alpha_i^*\right) K\left(\mathbf{x}, \mathbf{x}_i\right) + b_{\text{opt}} \tag{9}$$

where, $\alpha_i, \alpha_i^*$ are the Lagrange multipliers ($i$=1, 2,…, $N_s$), $K(\mathbf{x}, \mathbf{x}_i)$ is the kernel function, $N_s$ is the number of support vectors and $b_{\text{opt}}$ is the optimal bias. Usually, a Radial Basis Function (RBF) kernel, given by $K(\mathbf{x}, \mathbf{y})=\exp(-\|\mathbf{x}-\mathbf{y}\|^2 / 2\sigma^2)$ ($\sigma$ is a tuning parameter) is preferred [22,23]. The proper selection of the model parameters $C$ and $\sigma$ is crucial for effectiveness of SVM. Following other works [22,23], grid search and cross validation techniques are used in this paper for finding optimal SVM parameters.

## 5. EXPERIMENTS AND DISCUSSIONS

To empirically examine the effectiveness of our proposed ensemble technique, four widely popular real world time series from different domains are used. These are—the Canadian lynx, the Wolf's sunspot, the monthly international airline passengers, and the monthly Australian sales of red wine time series. All these four datasets are collected from the Time Series Data Library (TSDL) repository [20]. Table 1 gives the necessary descriptions about them and Fig. 2 shows the corresponding time plots.

**Table 1. Descriptions of the four time series datasets**

| Time Series | Description | Dataset Size |
|---|---|---|
| Lynx | Number of lynx trapped per year in the Mackenzie River district of Northern Canada (1821–1934). | Total size: 114<br>Training: 80<br>Validation: 20<br>Testing: 14 |
| Sunspots | The annual number of observed sunspots (1700–1987). | Total size: 288<br>Training: 171<br>Validation: 50<br>Testing: 67 |
| Airline Passengers | Monthly total number of international airline passengers (in thousands) (January 1949–December 1960). | Total size: 144<br>Training: 120<br>Validation: 12<br>Testing: 12 |
| Red Wine | Monthly Australian sales of red wine (thousands of liters) (January 1980–December 1995). | Total size: 187<br>Training: 144<br>Validation: 24<br>Testing: 19 |

The selection of a proper validation set is crucial for the success of our ensemble scheme. Here, we choose the lengths of the validation datasets in such a way that they approximately match with the lengths of the corresponding test sets for the four time series. All experiments in this paper are implemented through MATLAB. The appropriate ANN structures are determined on the basis of common model selection criteria, as discussed in details by Zhang et al. [4]. The suitable ANN model for each dataset is trained for 2000 epochs with every training algorithm. The PSO toolbox, developed by Birge [24] is used for implementing PSO-Trelea1 and PSO-Trelea2. Following previous studies [14], the number of swarm particles is chosen from the range of 24 to 30.

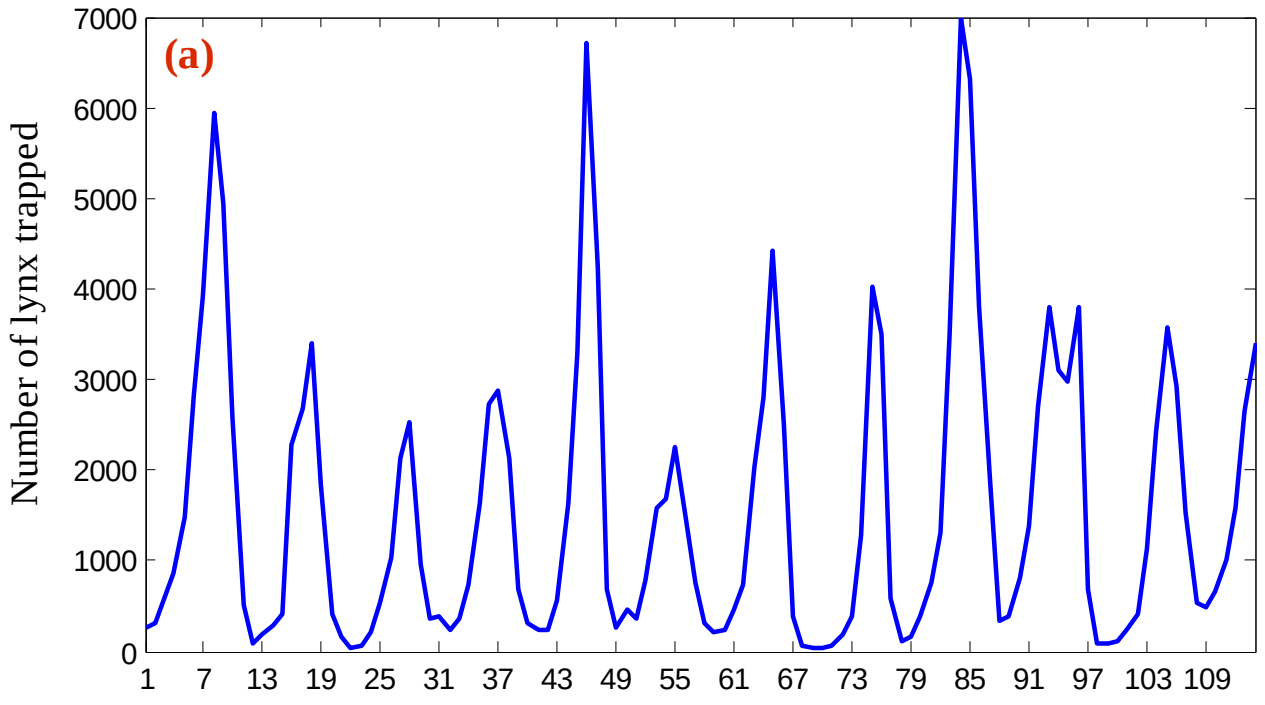

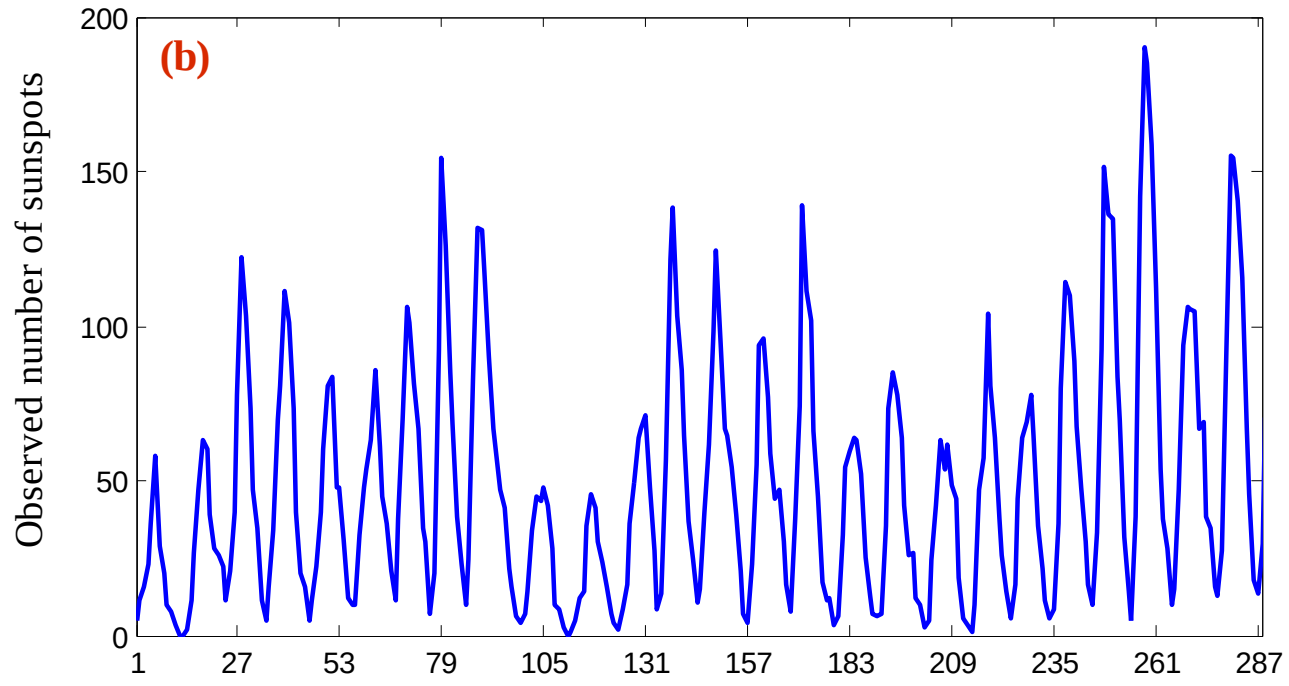

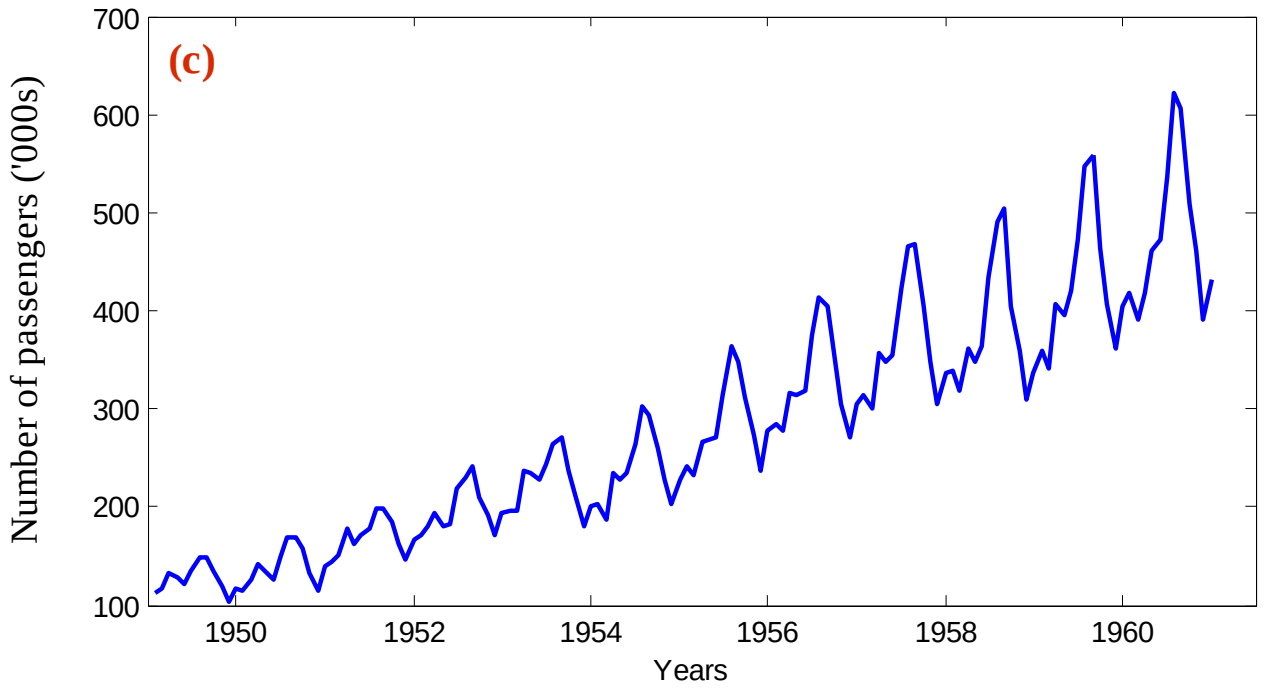

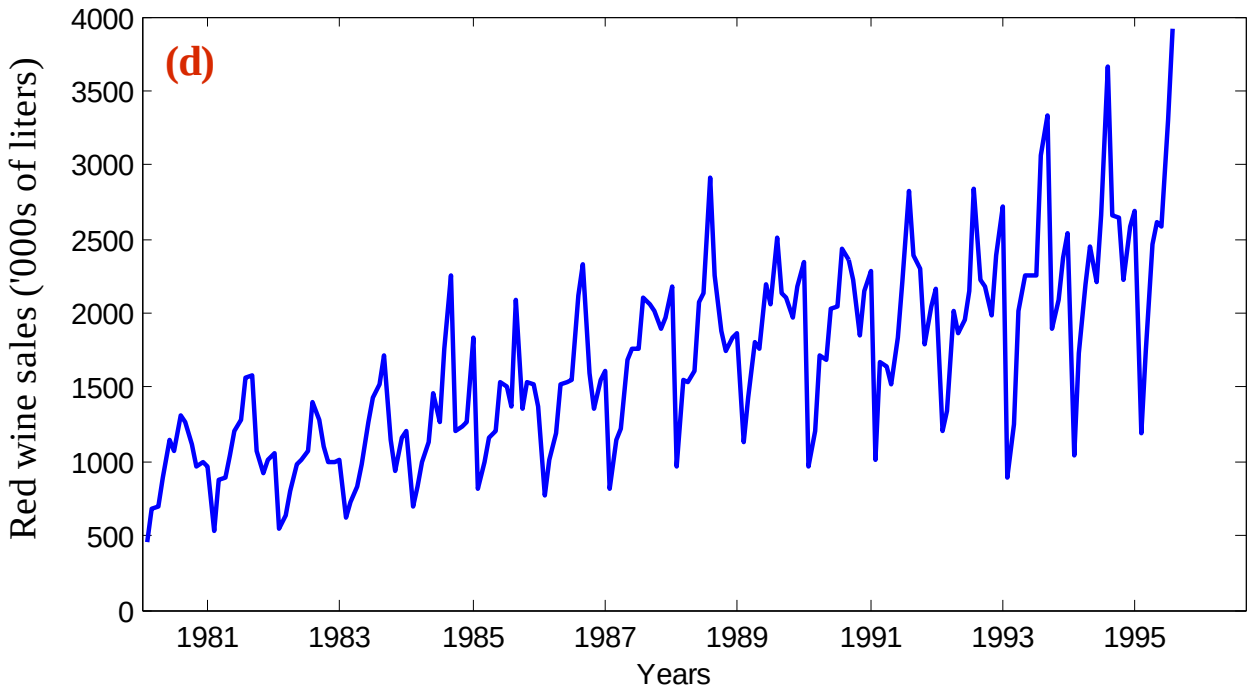

**Fig 2: Time series plots: (a) Canadian lynx, (b) Wolf's sunspots, (c) Airline passengers, (d) Red wine sales.**

The lynx and sunspots datasets are stationary and exhibit regular patterns. In particular, the sunspot series has a cycle of length approximately 11 years. In this study, the (7, 5, 1) and (4, 4, 1) ANN structures are found to be most suitable for lynx and sunspot series, respectively. Our findings agree with those by other works regarding these two time series [2]. Also, the ARIMA(12, 0, 0) (i.e. AR(12)) model, as employed by Hipel and McLeod is used for the lynx data and the ARIMA(9, 0, 0) (i.e. AR(9)) model [2] is used for the sunspot data. As suggested by Zhang [2], the logarithms (to the base 10) of the lynx data are used in the present analysis.

The airline passengers and red wine sales are nonstationary series, having monthly seasonal fluctuations (i.e. $s = 12$) with upward trends, as can be seen from Fig. 2(c) and 2(d). The seasonality in both series is strong and of multiplicative nature. The airline passenger series has been used by many researchers [1,26] for modeling trend and seasonal effect and now it is considered as a benchmark for seasonal datasets. Box and Jenkins were the first to determine that SARIMA$(0,1,1)\times(0,1,1)^{12}$ is the best stochastic model for the airline passenger series [1]. The same model is used in this paper for the airline data and incidentally it is found suitable for the red wine dataset too.

For ANN modeling of these two series, the Seasonal ANN (SANN) structure, developed by Hamzacebi [26] is considered in this paper. The unique characteristic of this model is that it uses the seasonal component of a time series to determine the number of input and output nodes. Also it can directly track the seasonal effect in the data without removing it. For a seasonal time series with period *s*, the SANN assumes a (*s*, *h*, *s*) ANN structure, *h* being the number of hidden nodes. This model is quite simple to understand and implement, yet very efficient in modeling seasonal data, as shown by the research work [26].

To evaluate the forecasting performances of all the fitted models, the two error measures, viz. MSE and MAPE are considered. The ANN model for each time series with every training algorithm is executed 50 times with different initial values for network weights and biases. The final errors are chosen as the best among these 50 runs. The boxplot in Fig. 3 presents a concise diagrammatic depiction of the relative reduction in MAPE (all four time series combined), achieved through our proposed ensemble technique. A similar result is also observed in case of the obtained forecast MSE values.

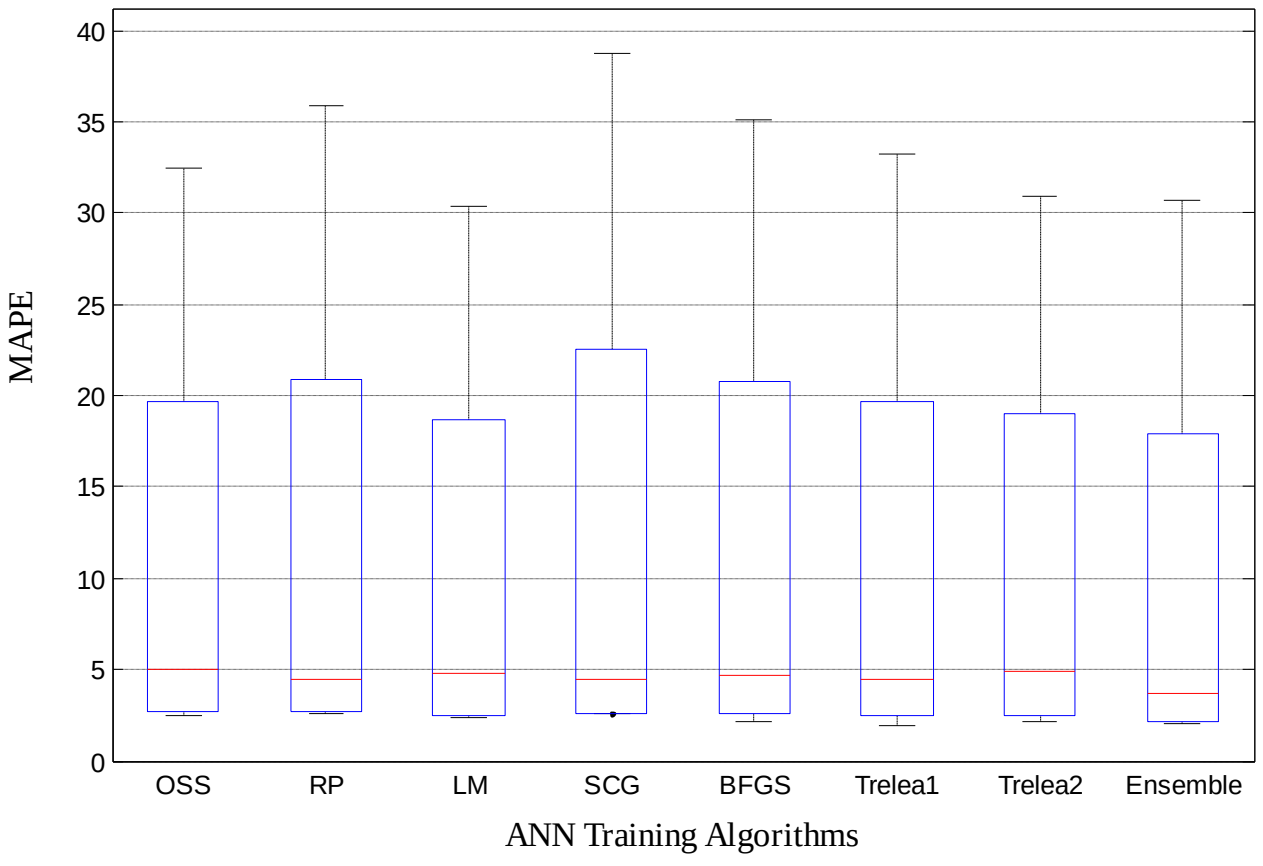

**Fig 3: Boxplot of the effect on forecast MAPE due to applying different ANN training algorithms**

The summary of the obtained error measures for all four datasets is presented in Table 2. In particular, the errors for airline and red wine datasets are given in transformed scales (obtained MSE=original MSE $\times 10^{-4}$).

**Table 2. Forecast comparison for all four time series**

| Models→ Forecast Errors↓ | | ARIMA | SVM | Proposed ANN Ensemble |
|---|---|---|---|---|
| Lynx | MSE | 0.01285 | 0.05267 | **0.00715** |
| | MAPE | 3.27743 | 5.81181 | **2.07280** |
| Sunspot | MSE | 483.491 | 792.961 | **280.478** |
| | MAPE | 60.0385 | 40.4331 | **30.6866** |
| Airline Passengers | MSE | 0.04118 | 0.01769 | **0.01485** |
| | MAPE | 3.70950 | 2.33661 | **2.16681** |
| Red Wine | MSE | 9.33719 | 12.8496 | **3.21148** |
| | MAPE | 9.64908 | 12.8592 | **5.17602** |

From Table 2, it is evident that the forecast errors obtained by our proposed ANN ensemble technique for all four datasets are much less than those obtained by ARIMA and SVM models. These empirical findings strongly support the fact that by combining multiple training algorithms the forecasting accuracy of an ANN model can be significantly improved. In this paper we use the term *Forecast Diagram* to refer the graph which shows the actual and forecasted observations for a time series. The obtained forecast diagrams for all four datasets are presented in Fig. 4.

## 6. CONCLUSIONS

During the last two decades ANNs have been extensively used for many time series forecasting problems. The wide popularity of ANNs in forecasting community can be credited to their many distinctive and excellent characteristics. However, the standard backpropagation network training method often suffers from a number of inherent drawbacks, such as: the complex pattern of error surfaces, slow convergence rates, getting stuck at local minima, etc. Although various improvements of the basic backpropagation technique have been developed in literature, but none of them could overcome all its shortcomings. Moreover, at present there is no rigorous way to select a best training algorithm specific to a particular problem.

In view of these facts, a novel weighted ensemble technique for combining multiple ANN training algorithms is proposed in this paper. The assignment of weights is based on the forecast performance of a training algorithm on the validation dataset. In this paper, seven different training algorithms are used for combining. The experiments conducted on four real world time series suggest that the ANN forecasting accuracies are significantly improved through this ensemble method. Moreover, it is also observed that this combined training algorithm performs much better than each of the individual ones. In future, the effectiveness of our proposed scheme can be further examined for other varieties of time series forecasting problems and with other model combination techniques.

## 7. ACKNOWLEDGMENTS

The first author would like to thank the Council of Scientific and Industrial Research (CSIR) for the obtained financial assistance, which helped a lot while performing this research.

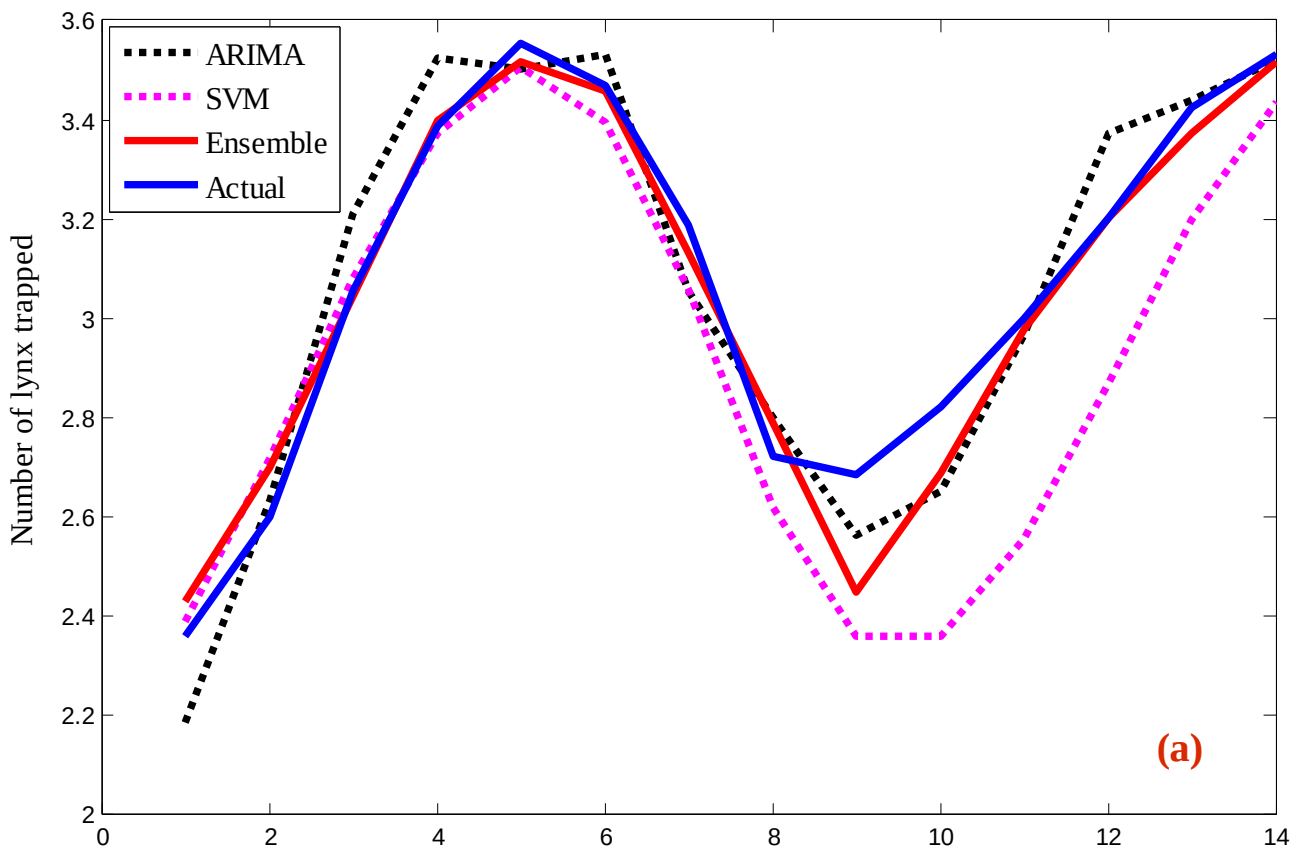

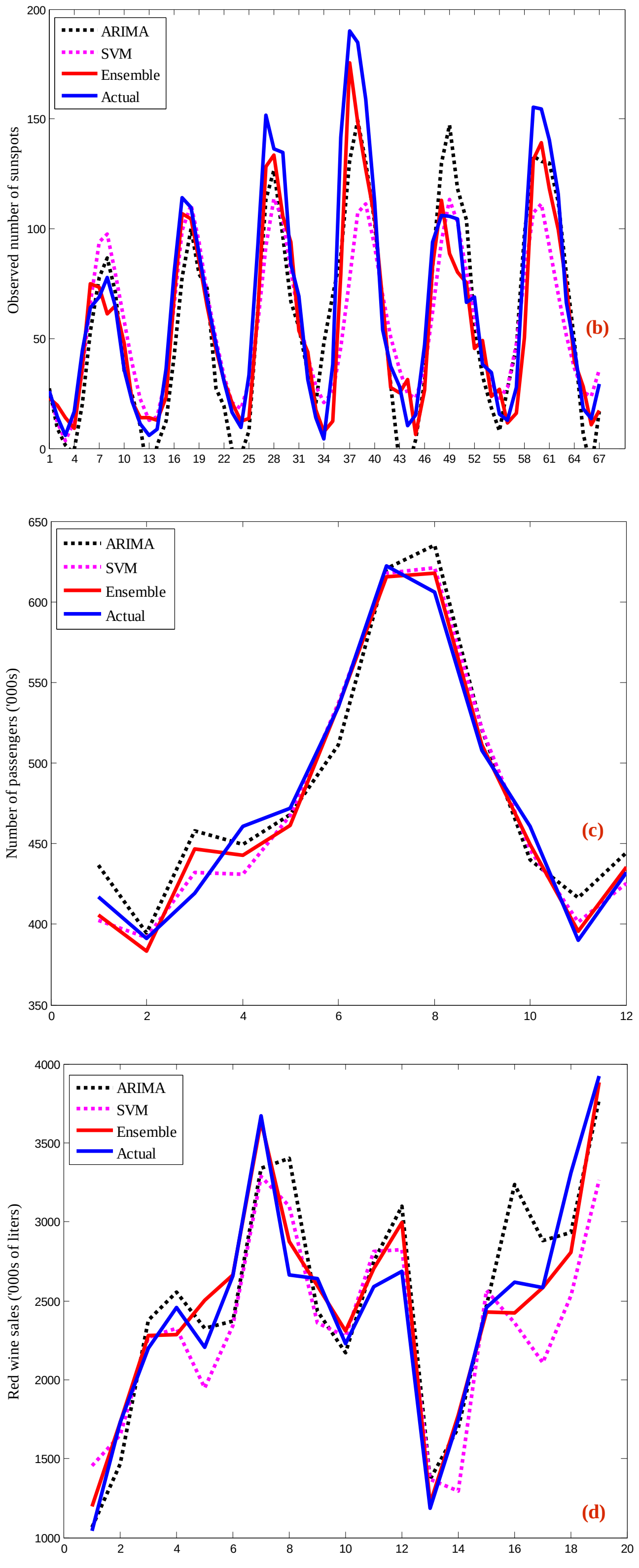

**Fig 4: Forecast diagrams: (a) Canadian lynx, (b) Wolf's Sunspots, (c) Airline passengers, (d) Red wine sales.**